\documentclass{ecai} 

\usepackage{latexsym}
\usepackage{amssymb}
\usepackage{amsmath}
\usepackage{amsthm}
\usepackage{booktabs}
\usepackage{enumitem}
\usepackage{graphicx}
\usepackage{color}

\usepackage{bm}
\usepackage{multirow}
\usepackage{color}
\usepackage{xcolor}
\usepackage{colortbl}
\usepackage{graphicx}
\usepackage{subcaption}
\usepackage{caption}
\usepackage{hyperref}

\newcommand{\BibTeX}{B\kern-.05em{\sc i\kern-.025em b}\kern-.08em\TeX}

\begin{document}


\begin{frontmatter}


\paperid{4194} 


\title{Generating Clinically Realistic EHR Data via a Hierarchy- and Semantics-Guided Transformer}


\author[A]{\fnms{Guanglin}~\snm{Zhou}\thanks{Corresponding Author. Email: guanglin.zhou@uq.edu.au}}
\author[A]{\fnms{Sebastiano}~\snm{Barbieri}}

\address[A]{University of Queensland}


\begin{abstract}
Generating realistic synthetic electronic health records (EHRs) holds tremendous promise for accelerating healthcare research, facilitating AI model development and enhancing patient privacy. However, existing generative methods typically treat EHRs as flat sequences of discrete medical codes. This approach overlooks two critical aspects: the inherent hierarchical organization of clinical coding systems and the rich semantic context provided by code descriptions. Consequently, synthetic patient sequences often lack high clinical fidelity and have limited utility in downstream clinical tasks. In this paper, we propose the Hierarchy- and Semantics-Guided Transformer (HiSGT), a novel framework that leverages both hierarchical and semantic information for the generative process. HiSGT constructs a hierarchical graph to encode parent-child and sibling relationships among clinical codes and employs a graph neural network to derive hierarchy-aware embeddings. These are then fused with semantic embeddings extracted from a pre-trained clinical language model (e.g., ClinicalBERT), enabling the Transformer-based generator to more accurately model the nuanced clinical patterns inherent in real EHRs. Extensive experiments on the MIMIC-III and MIMIC-IV datasets demonstrate that HiSGT significantly improves the statistical alignment of synthetic data with real patient records, as well as supports robust downstream applications such as chronic disease classification.
The code is available at \href{https://github.com/jameszhou-gl/HiSGT}{https://github.com/jameszhou-gl/HiSGT}.
.

\end{abstract}

\end{frontmatter}


\section{Introduction}

Electronic Health Records (EHRs) are a cornerstone of modern healthcare, playing a pivotal role in patient care, clinical decision-making, and medical research~\citep{jha2009use,cowie2017electronic,xiao2021introduction,xiao2018opportunities}. 
Despite their importance, strict regulations surrounding privacy and data confidentiality have made it challenging to share and utilize EHR data at scale~\citep{johnson2016mimic,johnson2023mimic}. 
As a result, there is growing interest in \textbf{synthetic EHR data generation}, a paradigm that aims to produce clinically realistic datasets mimicking the distributional properties of real EHRs while protecting sensitive patient information~\citep{gonzales2023synthetic,mcduff2023synthetic,van2024synthetic,ghosheh2024survey}. 

Recent advances in generative models have spurred significant progress in synthetic EHR data generation.
Many approaches use Generative Adversarial Networks (GANs)~\citep{choi2017generating,zhang2021synteg,torfi2020corgan,cui2020conan,zhang2020ensuring,rashidian2020smooth,kuo2022health} and Variational AutoEncoders (VAEs)~\citep{biswal2021eva}.
Other methods have drawn on the Transformer architecture~\citep{Vaswani2017AttentionIA} to capture long-range dependencies in clinical sequences.
In these models, medical codes and clinical events are typically treated as discrete tokens, with a next-token prediction objective used to model the sequential structure of EHR data~\citep{theodorou2023synthesize,kraljevic2024foresight,pang2024cehr,renc2024zero}.
However, they overlook the rich semantic information embedded in clinical code descriptions and the inherent hierarchical relationships in clinical coding systems.
Consequently, the synthesized patient sequences frequently lack high clinical fidelity, which limits the utility of the generated EHR data.
In particular, two major research gaps remain.

First, medical coding systems such as the International Classification of Diseases (ICD)~\citep{world1978international} are organized hierarchically, with codes exhibiting parent–child and sibling relationships that reflect key clinical knowledge. 
For example, in ICD-10, the code {`E11'} (Type 2 diabetes mellitus) serves as a parent for {`E11.6'} (type 2 diabetes mellitus with other specified complications) and as an ancestor for {`E11.65'} (Type 2 diabetes mellitus with hyperglycemia), while sibling codes such as {`E11.65'} and {`E11.64'} (Type 2 diabetes mellitus with hypoglycemia) capture different diabetes complications at a more granular level. 
In clinical practice, a patient may first be diagnosed at a coarse level with {`E11'} and later be assessed for specific complications (e.g., {`E11.64'} or {`E11.65'}).
\emph{However, current generative models often treat each code as an isolated symbol, failing to leverage these multi-level taxonomic properties}. 
As a result, they may generate conflicting sibling codes that cannot clinically co-occur (e.g., {`E11.64'} hypoglycemia and {`E11.65'} hyperglycemia in the same visit) or neglect important subcodes.

Second, each medical code is accompanied by a textual description that clarifies its clinical semantics and distinguishes between disease subtypes. 
For instance, {`E11.65'} denotes `Type 2 diabetes mellitus with hyperglycemia', while {`Z13.6'} specifies `screening for cardiovascular disorders'. 
Such text descriptions are essential for determining the precise clinical focus of a patient’s visit. 
In practice, each patient's visit to the hospital often centers on a primary diagnosis (or a small set of diagnoses) with additional procedures tailored to that theme until the patient's condition is confirmed by the physician.
Consequently, the codes assigned within a single visit tend to be semantically related, reflecting the shared focus on a particular disease process.
\emph{However, existing generative methods typically disregard these semantic descriptions and treat codes solely as discrete tokens}.
Without leveraging the underlying semantics, these models may synthesize codes that are logically or clinically incompatible.

\begin{figure*}[hbt]
\centering
\includegraphics[width=0.86\linewidth,keepaspectratio]{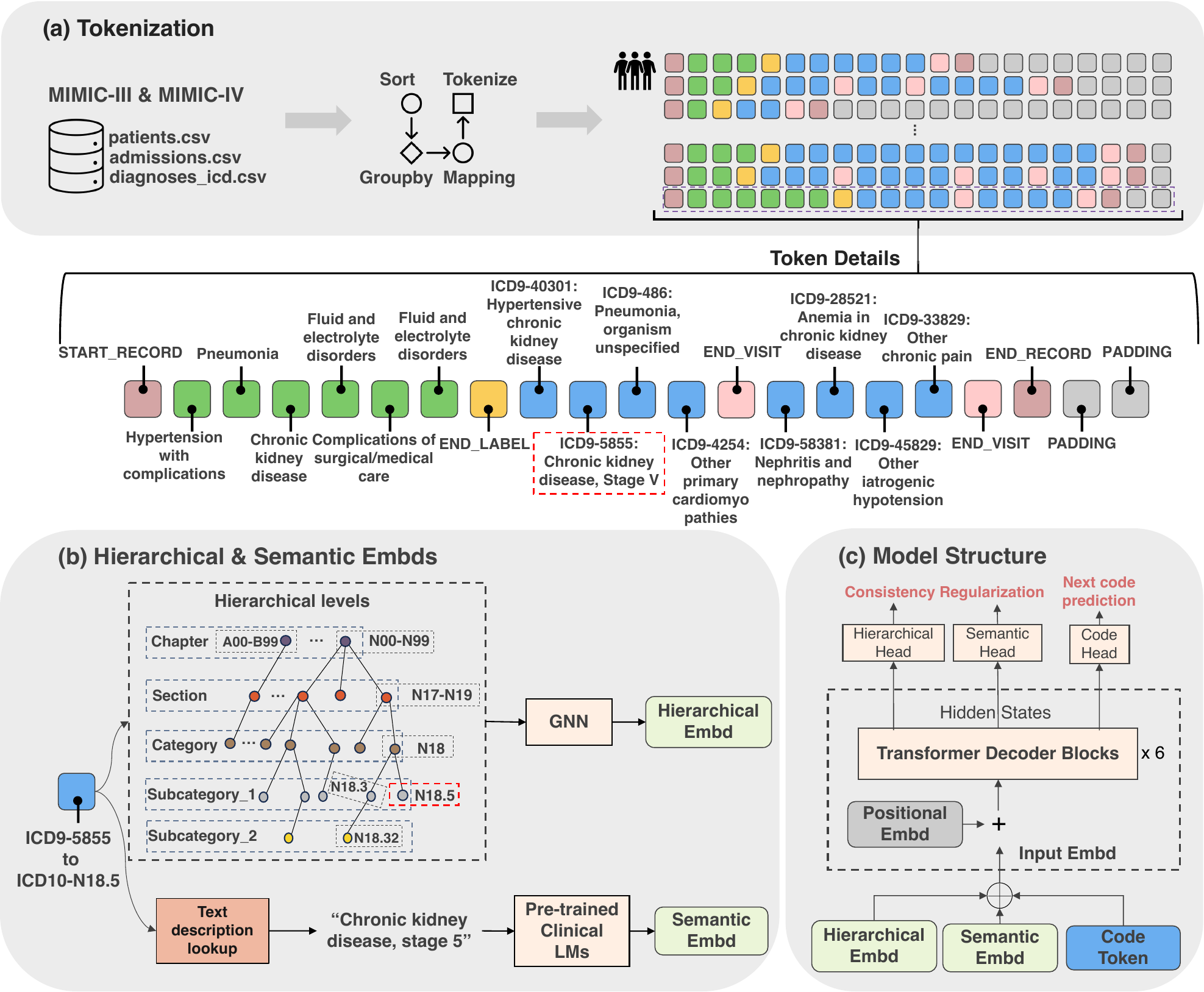}
\caption{\textbf{Overview of HiSGT for Synthetic EHR Generation.} \textbf{(a) Tokenization:} We extract structured patient records from the MIMIC-III and MIMIC-IV datasets, and process hospital admissions and diagnosis codes. The tokenization process encodes patient timelines into structured sequences, strictly following the format: \{\texttt{START\_RECORD}, phenotype label tokens, \texttt{END\_LABEL}, event tokens per visit, \texttt{END\_VISIT}, $\cdots$, \texttt{END\_RECORD}, \texttt{PADDING}\}. \textbf{(b) Hierarchical \& Semantic Embeddings:} Medical codes (e.g., `N18.5') are enriched with hierarchical embeddings from a constructed ICD-based graph and semantic embeddings from a pre-trained clinical language model (LM), e.g, ClinicalBERT. The GNN-derived hierarchical embeddings capture multi-level taxonomic relationships, while ClinicalBERT encodes text descriptions. \textbf{(c) Model Structure:} HiSGT integrates hierarchical and semantic embeddings into Transformer-based decoder blocks. Three prediction heads—hierarchical, semantic, and next-code prediction—jointly optimize the model. During inference, HiSGT autoregressively generates synthetic patient records, starting from the first token \texttt{START\_RECORD} and sequentially predicting phenotype labels, medical visit/events until reaching \texttt{END\_RECORD}.}
\label{213752426926}
\vspace{5pt}
\end{figure*}

To address these limitations, we propose a novel framework called Hierarchy and Semantics-Guided Transformer (HiSGT), which integrates both  hierarchical relationships and semantic information into the generative process. 
As shown in Figure~\ref{213752426926}, HiGST begins by constructing a hierarchical graph to mirror the taxonomy of real coding systems such as ICD, preserving the parent–child and sibling relationships. 
This graph helps to encode and extract clinically coherent paths for every single code. A graph neural network (GNN) is then employed to learn hierarchy-aware embeddings that capture structured dependencies among medical codes. Additionally, HiGST enriches each medical code with semantic embeddings derived from a pre-trained clinical language model (e.g. ClinicalBERT~\citep{alsentzer2019publicly}).
These embeddings are consistent with their text description to help inject essential domain knowledge into the Transformer backbone.
Thus, concepts like `Type 2 diabetes' carry meaningful distinctions rather than being treated solely as a meaningless token.
Finally, these refined embeddings are fed into a Transformer-based generative model to synthesize EHR data that are both semantically consistent and structurally coherent. 

Our work makes the following key contributions to the field of synthetic EHR generation: (1) we propose a novel framework that leverages both hierarchical graph representations and clinical semantic embeddings, as an approach to embed medical codes in general, to overcome the limitations of raw code-only generative methods; (2) we develop an enhanced Transformer model that integrates refined code representations, enabling the generation of synthetic EHRs with high clinical fidelity; (3) we demonstrate the effectiveness of HiSGT on two real-world EHR datasets, showing significant improvements in synthetic data fidelity, utility for downstream tasks (e.g., disease classification), and privacy preservation.

\section{Related Work}

\subsection{EHR Data Generation}
The generation of synthetic EHRs have evolved across several categories~\citep{chen2024generating}. Eearly rule-based methods rely on predefined clinical guidelines to simulate patient records~\citep{buczak2010data,mclachlan2018aten}. While these methods ensure clinically interpretability, they fail to capture complex correlations in real-world EHR data.
GAN-based approaches model the distribution of real EHR data through adversarial training. A generator synthesizes patient records, while a discriminator distinguishes between real and synthetic samples~\citep{zhang2020ensuring,yale2020generation,wang2024igamt,li2023generating,baowaly2019synthesizing,yang2019grouped,lee2020generating,yan2021generating,torfi2020corgan,kuo2022health}.
VAE-based methods learn a latent representation of EHR data and generate new samples by sampling from the learned distribution~\citep{biswal2021eva,sun2024collaborative}.
However, both of GANs and VAEs struggle with rare events in high-dimensional EHR data and capturing fine-grained longitudinal dependencies in EHR data.
Recently, Transformer-based generative models gained traction due to their ability to capture long-range dependencies in patient records.
HALO~\citep{theodorou2023synthesize} applies a self-attention mechanisms to model the sequential structure of medical codes, significantly expanding the vocabulary of synthetic data from fewer than 100 tokens to over thousands. However, existing methods rely solely on modeling raw code sequences.
As result, the generated records lack high clinical fidelity.
In this work, we propose a Transformer-based framework with two inductive biases of hierarchical and semantic information.

\subsection{Generative Transformer}
The Transformer architecture, introduced by ~\citep{Vaswani2017AttentionIA}, has revolutionized sequence modeling by effectively capturing long-range dependencies through self-attention mechanisms. 
In natural language processing (NLP), generative Transformer models such as GPTs series~\citep{radford2019language} have demonstrated SOTA performance in text generation. These models predict tokens sequentially while conditioning on all previous tokens, making them highly effective for learning complex dependencies. 
Beyond NLP, Transformers have been successfully adapted to other domains including computer vision~\citep{dosovitskiy2021imageworth16x16words} and multimodal learning~\citep{hurst2024gpt}.
In healthcare, Transformer-based models for synthetic EHR generation typically treat patient trajectories as sequences of discrete medical codes~\citep{theodorou2023synthesize, pang2024cehr}.
Although they enable scaling to synthetic data with thousands of unique tokens, they model medical codes purely as discrete symbols, thereby the generated EHR lack clinical fidelity.
Our work extends Transformer-based EHR generation by explicitly incorporating structured knowledge into the generative process—leveraging hierarchical representations and semantic embeddings inherent in clinical coding systems.

\section{Methodology}

\subsection{Problem Formulation}
\label{135320712381}
EHR data naturally exhibits a multi-level structure comprising \emph{patients}, \emph{visits} and \emph{clinical events}. Let $\mathcal{X}$, $\mathcal{V}$ and $\mathcal{E}$ denote the sets of patients, visits and clinical events, respectively. 
For each patient $\bm{x} \in \mathcal{X}$, the medical history is represented as a sequence of visits $\bm{x} = \{\bm{v}_1, \bm{v}_2, \ldots, \bm{v}_{n_x}\}$,
where $n_x$ is the number of visits. Each visit $\bm{v}_i$ comprises a set of medical codes representing diagnoses, procedures, and other clinical events:
$\bm{v}_i = \{e^i_1, e^i_2, \ldots, e^i_{m_i}\},$
with $m_i$ denoting the number of medical codes recorded during visit $\bm{v}_i$. 
The objective of synthetic EHR generation is to model the joint distribution of patient trajectories while preserving both intra-visit (within-visit) and inter-visit (longitudinal) dependencies. The model estimates the \emph{conditional probability} of each clinical event given past visits and previously observed events within the same visit:
\begin{equation}
\label{806165374009}
    {L}(\Theta) = \sum_{\bm{x} \in \mathcal{X}} \sum_{i=1}^{n_x} \sum_{j=1}^{m_i} \log P\Bigl(e^i_j \,\Big\vert\, \underbrace{\bm{v}_{1}, \cdots, \bm{v}_{i-1}}_{\text{inter-visit}}, \underbrace{e^i_1, \cdots, e^i_{j-1}}_{\text{intra-visit}};\, \Theta\Bigr)
\end{equation}
where $\Theta$ represents the model parameters.
The generative model seeks to maximize this likelihood to ensure that the generated sequences closely resemble real-world clinical patterns.
Once trained, this model generates realistic EHR trajectories auto-regressively. At each step $j$, the predicted event $e_j^i$ is sampled from the conditional distribution $P\Bigl(e^i_j \,\Big\vert\, \bm{v}_{1}, \cdots, \bm{v}_{i-1}, e^i_1, \cdots, e^i_{j-1}\Bigr)$, appended to the sequence, and used as input for the subsequent step. This process repeats until a predefined maximum length $L$ is reached or an `END\_RECORD' token is generated, signaling completion.

\subsection{HiSGT Framework}
HiSGT is a Transformer-based generative model designed for synthetic EHR data generation. To enhance clinical fidelity, the model incorporates hierarchical and semantic embeddings to enrich medical code representations, as illustrated in Figure~\ref{213752426926}.

\vspace{-3.5mm}
\subsubsection{Tokenization}
\label{995716973558}
The tokenization process in HiSGT structures raw EHR data into a sequence of discrete tokens while preserving temporal dependencies betweeen medical events. We extract longitudinal patient records from the MIMIC-III and MIMIC-IV datasets, using key tables such as \texttt{patients.csv}, \texttt{admissions.csv} and \texttt{diagnoses\_icd.csv}. Each patient's health timeline consists of one or more hospital visits, with each visit containing diagnosis codes that describe the patient's conditions. To ensure an accurate representation of patient trajectories, the raw data undergoes preprocessing, where hospital visits are first sorted chronologically to maintain the correct sequence of admissions. Diagnosis codes are then grouped by patient ID, ensuring that all events for a single patient remain together. Additionally, each patient is assigned phenotype labels derived from their diagnoses, according to predefined clinical mappings. Once preprocessed, the dataset is tokenized by constructing a vocabulary that includes all unique ICD codes, along with phenotype label tokens and special tokens. 

As shown in Figure~\ref{213752426926}(a), each patient sequence follows a structured format. It begins with a \texttt{START\_RECORD} token, marking the initialization of a new patient record. This is followed by phenotype label tokens that summarize the patient's conditions, such as \emph{Hypertension with complications} and \emph{Pneumonia and Chronic kidney disease}. These label tokens play a crucial role in conditioning the generation of subsequent visits, ensuring consistency with standard utility evaluations. The \texttt{END\_LABEL} token separates phenotype labels from individual hospital visits. Each visit is represented as a collection of medical event tokens corresponding to ICD codes, e.g., ICD9-40301 (Hypertensive chronic kidney disease) and ICD9-486 (Pneumonia, organism unspecified). To preserve temporal dependencies, events from the same visit are grouped and concluded with an \texttt{END\_VISIT} token. The sequence may progress through multiple visits, maintaining the correct ordering of hospital admissions. Finally, the \texttt{END\_RECORD} token indicates the end of a patient's medical history. Since patient trajectories vary in length, \texttt{PADDING} tokens are appended where necessary to ensure consistent input dimensions.

\vspace{-3.5mm}
\subsubsection{Hierarchical \& Semantic Embeddings}
HiSGT enriches medical code representations by incorporating hierarchical and semantic embeddings, as shown in Figure~\ref{213752426926}(b). These embeddings capture structured relationships among medical codes and encode clinical meanings beyond raw code-only identifiers. 

\textbf{Hierarchical Embeddings.} Medical coding systems such as ICD follow structured taxonomies with multiple levels of granularity, including chapter, section, category, subcategory\_1, subcategory\_2. For example, in ICD-10, the code `N18.32' is nested within broader categories such as `N18.3', `N18', `N17-N19' and `N00-N99'. 
To capture these relationships, we construct a hierarchical graph based on ICD-10 version \(\mathcal{G} = (\mathcal{V}, \mathcal{E})\),
where each node \(e \in \mathcal{V}\) represents an ICD-10 code and edges \(\bigl(u \to v\bigr)\in \mathcal{E}\) capture parent--child or sibling relationships extracted from the ICD ontology.
This graph is represented by an adjacency matrix 
\(\,A \in \{0,1\}^{|\mathcal{V}|\times|\mathcal{V}|}\),
with \(A_{u,v}=1\) if $u$ and $v$ share a hierarchical link. Each code $e$ is initially represented by an identity feature vector \(\bm{z}_e \in \mathbb{R}^{|\mathcal{V}|}\). 
A graph neural network (GNN) propagates information across this hierarchy to produce clinically meaningful embeddings:
$\bm{h}_e 
\;=\; 
\mathrm{GNN}(\bm{z}_e, \mathcal{G})$, where $\bm{h}_e$ is the learned hierarchical embedding for each code $e$. The GNN is trained using adjacency reconstruction, ensuring that embeddings for related medical codes remain close in the latent space: $\hat{A}_{u,v} = \sigma\bigl(\bm{h}_u^\top \bm{h}_v\bigr)$;
where \(\sigma(\cdot)\) is a sigmoid activation function. 
The reconstruction loss, defined as \(\mathcal{L}_{\mathrm{recon}}=\|\hat{A}-A\|\) with mean-squared error, encourages structurally related codes  (e.g., {`N18'} and {`N18.5'}) to have similar hierarchical embeddings, while unrelated codes remain distinct. 
The resulting hierarchical embedding $\bm{Z}_h$ is then incorporated into the final input representation.

\textbf{Semantic Embeddings.} Beyond hierarchical relationships, medical codes are accompanied by textual descriptions that provide additional clinical context. These descriptions are essential for distinguishing between related conditions, procedures, and medications. For example, the ICD-10 code {`E11.65'} specifies `Type 2 diabetes mellitus with hyperglycemia'  whereas {`Z13.6'} specifies `screening for cardiovasscular disorders'. Ignoring these textual descriptions can lead to synthetically generated records that fail to reflect meaningful clinical distinctions.
HiSGT employs ClinicalBERT~\citep{alsentzer2019publicly}, a pre-trained clinical language model, to encode these descriptions into dense semantic representations. Given the text description $T(e)$ corresponding to medical code $e$, ClinicalBERT generates a semantic embedding: $\bm{s}_e = \mathrm{BERT}(T(e))$. Since ClinicalBERT outputs embeddings in a fix and high-dimensional space, we apply a linear transformation to project them into the model's latent space: $\bm{Z}_s = \bm{W}_s \bm{s}_e + \bm{b}_s$; where $\bm{W}_s \in \mathbb{R}^{d_s \times d}$ and $\bm{b}_s \in \mathbb{R}^{d}$ are learnable parameters.

\textbf{Final Input Representation.} The final input representation $\bm{Z}$ integrates three components:
\begin{equation}
\label{562463336330}
    \bm{Z}
    \;=\;
    \bm{Z}_{t}
    \;+\;
    \bm{Z}_{h}
    \;+\;
    \bm{Z}_{s}
\end{equation}
Here, $\bm{Z}_t$ is the standard token embedding obtained from the embedding layer in Transformer. $\bm{Z}_h$ is the hierarchical embedding, derived from the GNN trained to capture relationships among medical codes. $\bm{Z}_s$ is the semantic embedding, extracted from the ClinicalBERT.
This enriched representation allows HiSGT to generate synthetic EHR sequences that are bot structure-aware and clinically meaningful.

\vspace{-3.5mm}
\subsubsection{Model Structure}
HiSGT employs a decoder-only Transformer architecture~\citep{radford2019language} for auto-regressive (AR) generation of EHR sequences. As shown in Figure~\ref{213752426926}(c), the input consists of standard token embedding $\bm{Z}_t$ with pre-computed hierarchical ($\bm{Z}_h$) and semantic ($\bm{Z}_s$) embeddings, plus positional embeddings.
This composite embedding is processed through six stacked Transformer decoder blocks using causal masking for realistic sequence generation. 
The final hidden stages are projected to three output heads: 1) a code head for next-token prediction via probabilities over the medical code vocabulary, 2) a hierarchical consistency head, and 3) s semantic consistency head.
The latter two heads act as regularization, enforcing alignment with the input hierarchical and semantic structures (derived from ICD taxonomy and ClinicalBERT descriptions, respectively) to ensure clinically meaningful and structured output. 

\vspace{-3.5mm}
\subsubsection{Training and Inference}
HiSGT training uses a multi-objective approach, balancing a standard \emph{cross-entropy loss} for next-token prediction with two auxiliary \emph{Mean Squared Error (MSE) losses} from the consistency heads. 
These auxiliary losses act as implicit regularization mechanisms, guiding the model toward generating sequences that maintain both structural coherence and semantic fidelity. 
For inference, HiSGT generates patient trajectories auto-regressively: starting with a \texttt{START\_RECORD} token, it predicts phenotype labels until \texttt{END\_LABEL}, then generates sequences of medical codes per visit until an \texttt{END\_RECORD} token is produced, completing the synthetic patient record.

\section{Experiments}
\begin{table*}[!tb]
\centering
\caption{\textbf{Performance on Fidelity Metrics.} 
The table reports the $R^2$ scores for four metrics—Unigram (marginal event distribution), Bigram (intra-visit co-occurrence), Seq Bigram (inter-visit transitions), and DimWise (dimension-wise correlation)—evaluated on both MIMIC-III and MIMIC-IV.
}
\label{017557960858}

\renewcommand{\arraystretch}{0.9} 

\resizebox{\textwidth}{!}{
\begin{tabular}{l | c c c c| c c c c}
\toprule
 \multirow{2}{*}{Methods} & \multicolumn{4}{c|}{\textbf{MIMIC-III}}  & \multicolumn{4}{c}{\textbf{MIMIC-IV}} \\ 
 & {Unigram} & {Bigram } & {Seq Bigram } & DimWise & {Unigram} & {Bigram } & {Seq Bigram } & DimWise\\
\midrule

LSTM & 0.939 & 0.522 & 0.339 & 0.806 & 0.970 & 0.736 & 0.737 & 0.822\\ 
GPT & 0.935 & \underline{0.858} & \underline{0.811} & \underline{0.940} & 0.969 & \underline{0.935} & 0.907 & \underline{0.957}\\ 
EVA  & \textbf{0.986} & 0.749 & 0.779 & 0.869 & - & - & - & -\\
SynTEG & 0.036 & -2.562 & -0.579 & -0.052 & 0.909 & 0.733 & 0.630  & 0.735\\
HALO-Coarse & 0.933 & 0.774 & 0.637 & 0.701 & 0.927 & 0.743 & 0.712 & 0.900\\
HALO & 0.936 & \underline{0.867} & \underline{0.869} & 0.764 & \underline{0.973} & 0.932 & \underline{0.924} & 0.947\\ 

\midrule
HiSGT (Ours) & \underline{0.984} & \textbf{0.949} & \textbf{0.879} & \textbf{0.976} & \textbf{0.989} & \textbf{0.967} &  \textbf{0.948} & \textbf{0.989}\\ 

\bottomrule
\end{tabular}}
\end{table*}

\begin{table*}[!htb]
\centering
\caption{
\textbf{Performance on Phenotype Classification.}
This table presents the results of the {Train on Synthetic, Test on Real} (\textbf{TSTR}) evaluation for phenotype classification. Models are trained on synthetic patient records and evaluated on real MIMIC-III and MIMIC-IV datasets. We report the average accuracy, precision, recall, and F1-score (with standard deviations) across 25 phenotype categories. 
}
\label{740713087553}
\resizebox{\textwidth}{!}{
\begin{tabular}{l | c c c c | c c c c}
\toprule
\multirow{2}{*}{Methods} & \multicolumn{4}{c|}{\textbf{MIMIC-III}} & \multicolumn{4}{c}{\textbf{MIMIC-IV}} \\
 & Acc &  Precision & Recall & F1-Score & Acc &  Precision & Recall & F1-Score \\
\midrule
\rowcolor{gray!20} Real Data &  0.951$\pm$0.02 &  0.946$\pm$0.02 &  0.957$\pm$0.03 &  0.951$\pm$0.02 &  0.945$\pm$ 0.02&  0.936$\pm$0.02 &  0.956$\pm$0.02 &  0.946$\pm$0.02 \\ \midrule

LSTM &  0.493$\pm$0.06 &  0.480$\pm$0.12 &  0.673$\pm$0.26 &  0.538$\pm$0.17 &  0.545$\pm$0.10 &  0.534$\pm$0.15 &  0.580$\pm$0.18 &  0.546$\pm$0.14   \\ 
GPT  &  0.885$\pm$0.04 &  0.862$\pm$0.05 &  0.919$\pm$0.05 &  0.889$\pm$0.04 &  0.883$\pm$0.05 &  0.864$\pm$0.05 &  0.912$\pm$0.05 &  0.887$\pm$0.04  \\ 
EVA &  0.500$\pm$0.07 &  0.470$\pm$0.15 &  0.636$\pm$0.26 &  0.526$\pm$0.17 &  - &  - &  - &  -  \\
SynTEG  &  0.515$\pm$0.03 &  0.605$\pm$0.17 &  0.680$\pm$0.38 &  0.513$\pm$0.22 & 
0.563$\pm$0.08 &  0.622$\pm$0.11 &  0.558$\pm$0.26 &  0.533$\pm$0.12  \\
HALO-Coarse &  0.858$\pm$0.05 &  0.855$\pm$0.05 &  0.865$\pm$0.07 &  0.859$\pm$0.05 & 
0.827$\pm$0.05 &  0.827$\pm$0.05 &  0.830$\pm$0.08 &  0.826$\pm$0.05\\
HALO  &  0.892$\pm$0.03 &  0.871$\pm$0.04 &  0.920$\pm$0.04 &  0.895$\pm$0.03 &  0.893$\pm$0.04 &  0.880$\pm$0.04 &  0.909$\pm$0.05 &  0.894$\pm$0.04   \\ 
\midrule

HiSGT (Ours) &  \textbf{0.898}$\pm$0.04 &  \textbf{0.878}$\pm$0.05 &  \textbf{0.927}$\pm$0.05 &  \textbf{0.901}$\pm$0.04 &  \textbf{0.905}$\pm$0.04 &  \textbf{0.888}$\pm$0.03 &  \textbf{0.929}$\pm$0.04 &  \textbf{0.907}$\pm$0.04   \\
\bottomrule
\end{tabular}}
\end{table*}

\subsection{Dataset and Baseline Details}
We conduct experiments on two widely used real-world EHR datasets: \textbf{MIMIC-III}~\citep{johnson2016mimic} and \textbf{MIMIC-IV}~\citep{johnson2023mimic}.
We follow the tokenization steps in \S \ref{995716973558}.
After preprocessing, MIMIC-III consists of 46,520 patients with 6,984 unique ICD diagnosis codes and 7,012 unique tokens, while MIMIC-IV includes 124,525 patients with 9,072 unique ICD codes and 9,102 unique tokens.
Both datasets contain structured longitudinal patient records with multiple hospital visits, making them well-suited for evaluating synthetic EHR generation models.
To assess the effectiveness of HiSGT, we compare it against six state-of-the-art (SOTA) generative models used for EHR synthesis. LSTM~\citep{lee2018natural} is a recurrent neural network that models sequential dependencies. GPT~\citep{radford2019language} applies a transformer-based autoregressive approach for next-token prediction. EVA~\citep{biswal2021eva} uses a variational autoencoder (VAE) to learn latent representations of EHR sequences. SynTEG~\citep{zhang2021synteg} employs a GAN-based framework to generate synthetic patient visits. HALO-Coarse~\citep{theodorou2023synthesize} treats each visit as a single token, modeling longitudinal dependencies at a coarse level, while HALO~\citep{theodorou2023synthesize} refines this approach by representing visits at a more granular level.

\vspace{-3.5mm}
\subsection{Experimental Setting}
\label{36255018}
We conduct experiments using the MIMIC-III v1.4 and MIMIC-IV v2.2 datasets.  
The data is split into 80\%-20\% for training and testing, with an additional 90\%-10\% split within the training set for validation.  
For HiSGT, we use a six-layer Transformer architecture with eight attention heads and a hidden size of 384.  
Training is performed using the Adam optimizer with a learning rate of $10^{-4}$, a batch size of 48, and a dropout rate of 0.1.  
The input sequence length is set to 768 tokens, and training is run for 100 epochs with early stopping (patience = 10) based on validation loss.  
The best model checkpoint is saved according to minimum validation loss.  
The model is implemented in Python 3.13.0 using PyTorch 2.5.0, along with scikit-learn 1.5.2, NumPy 2.1.2, and transformers 4.46.2.  
All experiments are conducted on a single NVIDIA H100 GPU (80GB VRAM).  
\vspace{-3.5mm}
\subsection{Main Results}

\subsubsection{Fidelity Evaluation}

To assess the clinical fidelity of synthetic EHR data, we employ four complementary metrics that assess the alignment between real and synthetic records using the $R^2$ coefficient. These capture individual event distributions and relationships within and across visits.
First, the \textbf{Unigram} score measures how well the marginal distribution of individual medical events is preserved by comparing event frequencies. 
Beyond individual events, the \textbf{Bigram} score evaluates intra-visit coherence by capturing statistical dependencies between pairs of medical events within the same visit—an essential property, as these events often shown strong correlations. 
Similarly, the (\textbf{Seq Bigram}) score extends this to \emph{inter-visit} dependencies, quantifying how well longitudinal patterns (transitions between medical codes across consecutive visits) are captured. Finally, (\textbf{DimWise}) evaluates alignment at the patient level by comparing the mean per-patient probability distribution of medical codes. We compute the relative frequency of each code per patient, average these probabilities across all patients for each dataset (real and synthetic), and calculate the $R^2$ between the resulting vectors.

The results in Table~\ref{017557960858} provide evidence of HiSGT’s superior ability to synthesize clinically faithful EHR data. For the MIMIC-III dataset, HiSGT achieves an Unigram score of 0.984, closely approximating the real data’s marginal distribution. More importantly, the intra-visit Bigram score reaches 0.949—an 8.2\% improvement over HALO’s 0.867—indicating that HiSGT more accurately captures the co-occurrence patterns of medical events within a visit. The sequential dependencies, as measured by the Seq Bigram score, are also better modeled by HiSGT (0.879 versus 0.869 for HALO), while the DimWise metric (0.976) further confirms its effectiveness in preserving dimension-level correlations. Similarly, on the MIMIC-IV dataset, HiSGT consistently outperforms competing methods by recording a Unigram score of 0.989, a Bigram score of 0.967, and a Seq Bigram score of 0.948, compared to HALO’s scores of 0.973, 0.932, and 0.924, respectively. The DimWise score for HiSGT reaches 0.989, reinforcing its capability to capture both the marginal distributions and the complex dependencies inherent in clinical data. These improvements underscore the importance of incorporating hierarchical and semantic structures in generating synthetic EHR data.

\vspace{-3.5mm}
\subsubsection{Utility Evaluation}

\begin{table*}[!htb]
\centering
\caption{\textbf{Phenotype Classification Performance Across Disease Categories.} This table provides a detailed breakdown of phenotype classification performance for different disease types (acute, chronic, and mixed). We compare models trained on synthetic data (HALO and HiSGT) with real data as the ground truth. 
Accuracy and F1-scores are reported for each phenotype category, with macro-averaged scores included for all acute, mixed, and chronic disease groups. 
}
\label{421741622055}

\renewcommand{\arraystretch}{0.9} 

\resizebox{\textwidth}{!}{
\begin{tabular}{l | l | c c| c c | >{\columncolor{gray!20}}c >{\columncolor{gray!20}}c}
\toprule
\multirow{2}{*}{\textbf{Phenotype}} & \multirow{2}{*}{\textbf{Type}} & \multicolumn{2}{c|}{\textbf{HALO (SOTA)}} & \multicolumn{2}{c|}{\textbf{HiSGT (Ours)}} & \multicolumn{2}{c}{\cellcolor{gray!30}\textbf{Real Data}} \\
 & & Acc &   F1-Score  & Acc &   F1-Score   & Acc &   F1-Score  \\
\midrule
Acute and unspecified renal failure & acute & 0.917 & 0.919 & \textbf{0.921} & \textbf{0.924} & \cellcolor{gray!20} 0.960 & \cellcolor{gray!20} 0.960 \\
Acute cerebrovascular disease & acute  & 0.879 & 0.876 & \textbf{0.907} & \textbf{0.912} & \cellcolor{gray!20} 0.959 & \cellcolor{gray!20} 0.959 \\
Acute myocardial infarction & acute & 0.887 & 0.890 & \textbf{0.917} & \textbf{0.919} & \cellcolor{gray!20} 0.952 & \cellcolor{gray!20} 0.953 \\
Cardiac dysrhythmias & mixed & \textbf{0.898} & \textbf{0.899} & {0.893} & {0.895} & \cellcolor{gray!20} 0.937 & \cellcolor{gray!20} 0.938 \\
Chronic kidney disease& chronic  & \textbf{0.956} & \textbf{0.957} & {0.946} & {0.947} & \cellcolor{gray!20} 0.975 & \cellcolor{gray!20} 0.975 \\
Chronic obstructive pulmonary disease & chronic & \textbf{0.913} & \textbf{0.916} & {0.909} & {0.911} & \cellcolor{gray!20} 0.936 & \cellcolor{gray!20} 0.937 \\
Complications of surgical/medical care & acute  & {0.865} & {0.868} & \textbf{0.892} & \textbf{0.893} & \cellcolor{gray!20} 0.916 & \cellcolor{gray!20} 0.918 \\
Conduction disorders& mixed  & 0.878 & 0.881 & \textbf{0.879} & \textbf{0.885} & \cellcolor{gray!20} 0.937 & \cellcolor{gray!20} 0.938 \\
Congestive heart failure; nonhypertensive & mixed & {0.930} & {0.932} & \textbf{0.941} & \textbf{0.943} & \cellcolor{gray!20} 0.981 & \cellcolor{gray!20} 0.981 \\
Coronary atherosclerosis and other heart disease & chronic & \textbf{0.927} & \textbf{0.928} & {0.919} & {0.920} & \cellcolor{gray!20} 0.956 & \cellcolor{gray!20} 0.956 \\
Diabetes mellitus with complications & mixed & 0.916 & 0.916 & \textbf{0.919} & \textbf{0.922} & \cellcolor{gray!20} 0.956 & \cellcolor{gray!20} 0.956 \\
Diabetes mellitus without complication & chronic & 0.913 & 0.912 & \textbf{0.931} & \textbf{0.932} & \cellcolor{gray!20} 0.955 & \cellcolor{gray!20} 0.956 \\
Disorders of lipid metabolism & chronic & 0.916 & 0.917 & \textbf{0.923} & \textbf{0.926} & \cellcolor{gray!20} 0.963 & \cellcolor{gray!20} 0.963 \\
Essential hypertension & chronic & \textbf{0.946} & \textbf{0.946} & {0.939} & {0.940} & \cellcolor{gray!20} 0.957 & \cellcolor{gray!20} 0.958 \\
Fluid and electrolyte disorders & acute & 0.893 & 0.896 & \textbf{0.925} & \textbf{0.926} & \cellcolor{gray!20} 0.962 & \cellcolor{gray!20} 0.963 \\
Gastrointestinal hemorrhage & acute & 0.886 & 0.885 & \textbf{0.902} & \textbf{0.905} & \cellcolor{gray!20} 0.930 & \cellcolor{gray!20} 0.932 \\
Hypertension with complications & chronic & \textbf{0.941} & \textbf{0.941} & {0.939} & {0.940} & \cellcolor{gray!20} 0.977 & \cellcolor{gray!20} 0.977 \\
Other liver diseases & mixed & 0.875 & 0.877 & \textbf{0.892} & \textbf{0.891} & \cellcolor{gray!20} 0.946 & \cellcolor{gray!20} 0.947 \\
Other lower respiratory disease & acute & 0.842 & 0.848 & \textbf{0.855} & \textbf{0.850} & \cellcolor{gray!20} 0.952 & \cellcolor{gray!20} 0.952 \\
Other upper respiratory disease & acute & 0.749 & 0.723 & \textbf{0.767} & \textbf{0.770} & \cellcolor{gray!20} 0.923 & \cellcolor{gray!20} 0.924 \\
Pleurisy; pneumothorax; pulmonary collapse & acute & 0.883 & 0.887 & \textbf{0.884} & \textbf{0.888} & \cellcolor{gray!20} 0.949 & \cellcolor{gray!20} 0.949 \\
Pneumonia & acute  & 0.880 & 0.885 & \textbf{0.896} & \textbf{0.899} & \cellcolor{gray!20} 0.942 & \cellcolor{gray!20} 0.944 \\
Respiratory failure; insufficiency; arrest & acute  & 0.886 & 0.890 & \textbf{0.897} & \textbf{0.900} & \cellcolor{gray!20} 0.941 & \cellcolor{gray!20} 0.942 \\
Septicemia (except in labor)& acute  & 0.913 & 0.914 & \textbf{0.923} & \textbf{0.924} & \cellcolor{gray!20} 0.974 & \cellcolor{gray!20} 0.974 \\
Shock & acute & 0.900 & 0.899 & \textbf{0.921} & \textbf{0.924} & \cellcolor{gray!20} 0.965 & \cellcolor{gray!20} 0.965 \\
\midrule
All acute diseases (macro-averaged) &  & 0.893 & 0.894 & \textbf{0.905} & \textbf{0.907} & 0.945 & 0.946 \\
All mixed diseases (macro-averaged) &  & 0.892 & 0.893 & \textbf{0.901} & \textbf{0.903} & 0.943 & 0.944 \\
All chronic diseases (macro-averaged) &  & 0.923 & 0.925 & \textbf{0.932} & \textbf{0.933} & 0.963 & 0.964 \\
All diseases (macro-averaged) & & 0.893 & 0.894 & \textbf{0.905} & \textbf{0.907} & 0.945 & 0.946 \\
\bottomrule

\end{tabular}}
\end{table*}
\begin{table*}[!tb]
\centering
\caption{\textbf{Privacy Evaluation Results:} performance of the membership Inference Attack (MIA) and Attribute Inference Attack (AIA) on MIMIC-III and MIMIC-IV. 
Accuracy near 50\% and low AIA-F1 scores indicate better privacy. 
}

\label{852788858994}
\renewcommand{\arraystretch}{0.8} 
\resizebox{\textwidth}{!}{
\begin{tabular}{l | c c c c c| c c c c c}
\toprule
 \multirow{2}{*}{Methods} & \multicolumn{5}{c|}{\textbf{MIMIC-III}}  & \multicolumn{5}{c}{\textbf{MIMIC-IV}} \\ 
 & {Acc.} & {Precision} & {Recall} & F1-score & AIA-F1 & {Acc.} & {Precision} & {Recall} & F1-score & AIA-F1\\
\midrule
LSTM  & 0.500 & 0.500 & 0.494 & 0.497 & 0.008 & 0.507 & 0.507 & 0.502 & 0.505 & 0.011\\ 
GPT  & 0.504 & 0.504 & 0.462 & 0.482 & 0.036 & 0.511 & 0.512 & 0.480 & 0.495 & 0.044\\ 
EVA  & 0.500 & 0.500 & 0.462 & 0.480 & 0.007 & - & - & - & - & - \\
SynTEG  & 0.509 & 0.510 & 0.482 & 0.495 & 0.007 & 0.516 & 0.517 & 0.484 & 0.500 & 0.022\\
HALO-Coarse & 0.501 & 0.501 & 0.478 & 0.489 & 0.022 & 0.513 & 0.514 & 0.494 & 0.504 & 0.027\\
HALO  & 0.494 & 0.493 & 0.448 & 0.470 & 0.035 & 0.509 & 0.509 & 0.494 & 0.502 & 0.046\\ 
\midrule
HiSGT (Ours) & 0.493 & 0.492 & 0.454 & 0.472 & 0.034 & 0.504 & 0.504 & 0.478 & 0.491 & 0.045\\
\bottomrule
\end{tabular}}
\end{table*}

We evaluate clinical utility using a \textbf{Train on Synthetic, Test on Real (TSTR)} evaluation paradigm, where models are first trained on synthetic data and then evaluated on real patient outcomes~\citep{esteban2017real}.
We adopt the 25-phenotype classification benchmark~\citep{harutyunyan2019multitask}.
For real data, labels are assigned using a rule-based mapping from ICD-to-phenotype mapping (HCUP Clinical Classifications Software)~\citep{harutyunyan2019multitask}. Each patient in the real dataset is assigned one or more phenotype labels based on their recorded ICD codes. 
For synthetic data, labels are generated first, followed by conditional generation of patient visit sequences, ensuring meaningful label-event associations~\citep{esteban2017real}.
We employ a bi-directional LSTM classifier~\citep{theodorou2023synthesize}.
For each dataset (MIMIC-III/IV), we randomly extract 5,000 patient records for training, maintaining a balanced distribution of positive and negative labels for every single chronic disease category. Further, we reserve a separate validation set of 250 records for model selection and 500 real patient records as a held-out test set.

Table~\ref{740713087553} shows TSTR results. 
Models trained on real data serves as an upper bound. 
Among synthetic methods, HiSGT consistently achieves the highest performance, surpassing HALO and other baseline methods on both datasets. This demonstrates that HiSGT-generated patient trajectories retain the clinical coherence necessary for accurate phenotype classification.
Furthermore, detailed results across disease types in Table~\ref{421741622055} confirm HiSGT's superior generalization compared to SOTA method-HALO for acute, chronic, and mixed conditions.

\vspace{-3.5mm}
\subsubsection{Privacy Evaluation}

To ensure synthetic data do not compromise patient privacy~\citep{giuffre2023harnessing}, we evaluate protection against two common attacks: Membership Inference Attack (MIA) and Attribute Inference Attack (AIA).
The \textbf{MIA} tests if an adversary can identify records used during training. We follow a standard setup: randomly sample 500 records from the training set as \emph{members (positive)} and 500 from the test set (\emph{non-members}). For each record, we compute its minimum Hamming distance to the synthetic dataset and classify it as a member if its distance is below the median distance of the attack set. An accuracy near 50\% indicates privacy preservation close to random guessing.
The \textbf{AIA} assesses if sensitive attributes (medical codes) can be inferred from synthetic data. We aggregate visit-level information and use the 100 most frequent training codes as potential attributes. For each record, we predict undisclosed codes using a majority vote from its nearest neighbors (based on set distance) in the synthetic data. Attack performance is measured by F1-score (AIA-F1), with lower scores indicating better privacy.

Table~\ref{852788858994} summarizes the results. All evaluated models, including HiSGT, exhibit MIA accuracy close to 50\% and low AIA-F1 scores, confirming reasonable protection against these attacks.  
In particular, the results highlight an inherent privacy-utility trade-off: models like LSTM, EVA, and SynTEG achieve the strongest privacy scores (lowest AIA-F1, such as 0.007–0.008 on MIMIC-III) but suffer from poor fidelity and utility (as shown in Tables~\ref{017557960858} and~\ref{740713087553}).
Conversely, SOTA methods like HALO and our HiSGT deliver higher fidelity and utility. When compared to HALO, HiSGT achieves similar privacy protection with an MIA F1-score of 0.472 and an AIA-F1 of 0.034 on MIMIC-III, while continuing to improve fidelity and utility.

\begin{table*}[!htb]
\centering
\caption{\textbf{Ablation Study.} This table presents the impact of integrating hierarchical embeddings, semantic embeddings, and consistency constraints into the base Transformer model.
}
\label{219470308897}
\renewcommand{\arraystretch}{0.9} 
\resizebox{\textwidth}{!}{
\begin{tabular}{l | c c c c| c c c c}
\toprule
 \multirow{2}{*}{Methods} & \multicolumn{4}{c|}{\textbf{MIMIC-III}}  & \multicolumn{4}{c}{\textbf{MIMIC-IV}} \\ 
 & {Unigram} & {Bigram } & {Seq Bigram } & DimWise & {Unigram} & {Bigram } & {Seq Bigram } & DimWise\\
\midrule
Base (Code Only) & 0.956 & 0.898 & 0.821 & 0.964 & 0.965 & 0.929 & 0.910 & 0.975 \\ \midrule
\ + Hierarchical Embd & 0.967 & 0.914 & 0.845 & 0.972 & 0.983 & 0.950 & 0.941 & 0.988 \\
\ + Semantic Embd & 0.964 & 0.913 & 0.836 & 0.971 & 0.982 & 0.957 & 0.948 & 0.985\\ \midrule
\ + Hierarchical \& Semantic Embds & \multirow{1}{*}{0.976} & \multirow{1}{*}{0.927} & \multirow{1}{*}{0.859} & \multirow{1}{*}{\textbf{0.984}} & \multirow{1}{*}{0.987} & \multirow{1}{*}{0.960} & \multirow{1}{*}{\textbf{0.953}} & \multirow{1}{*}{\textbf{0.989}}\\ 
\midrule
\ + Hierarchical Embd \& Consistency & 0.973 & 0.926 & 0.844 & 0.967 & 0.984 & 0.949 & 0.933 & 0.986 \\
\ + Semantic Embd \& Consistency & 0.967 & 0.912 & 0.847 & 0.966 & 0.983 & 0.952 & 0.942 & 0.988\\
\midrule
\ Full HiSGT & \multirow{1}{*}{\textbf{0.984}} & \multirow{1}{*}{\textbf{0.949}} & \multirow{1}{*}{\textbf{0.879}} & \multirow{1}{*}{0.976} & \multirow{1}{*}{\textbf{0.989}} & \multirow{1}{*}{\textbf{0.967}} &  \multirow{1}{*}{0.948} & \multirow{1}{*}{\textbf{0.989}}\\ 
\bottomrule
\end{tabular}}
\end{table*}

\begin{figure*}[hbt]
\centering
\includegraphics[width=0.86\linewidth,keepaspectratio]{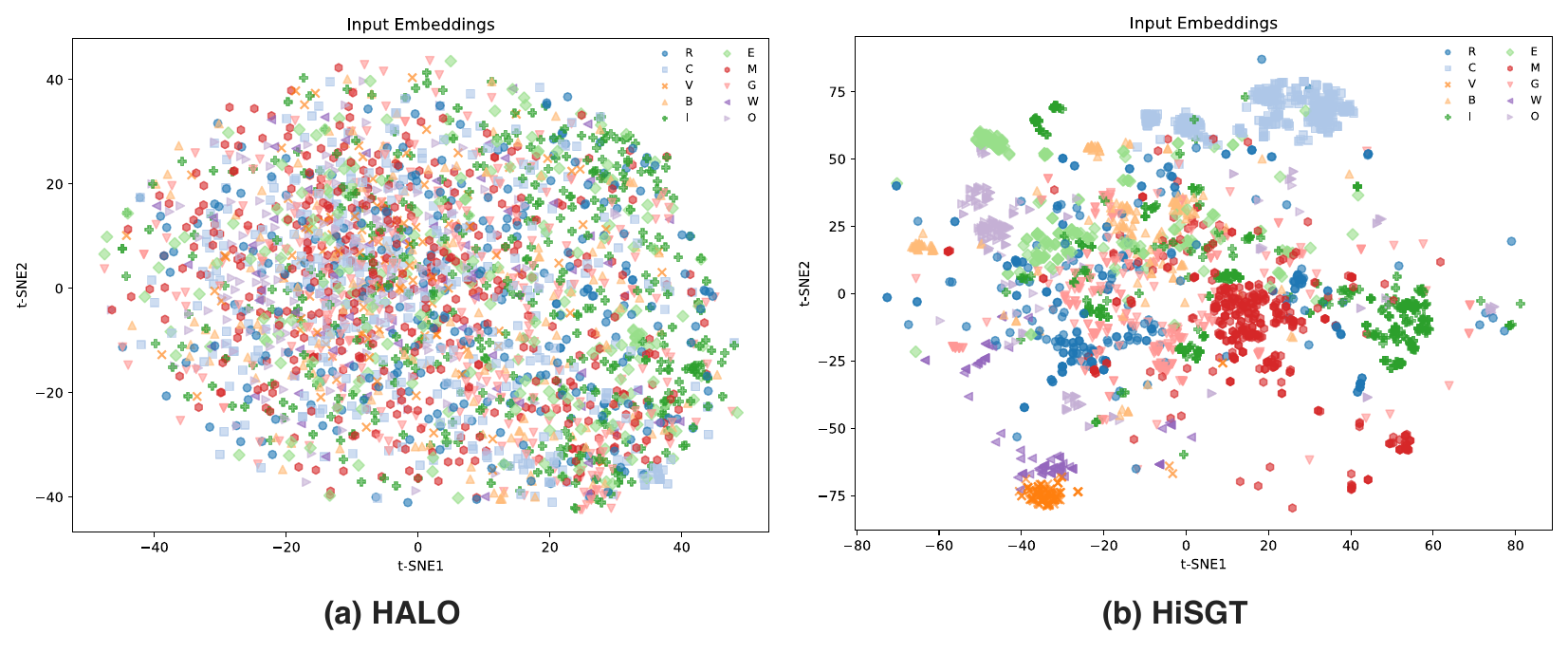}
\caption{\textbf{t-SNE visualizations of input embeddings.} Medical codes are plotted based on their input representations for (a) HALO and (b) HiSGT, colored by ICD category.
(a) HALO's standard embeddings show poor separation between categories, forming an undifferentiated cloud. (b) HiSGT's embeddings, incorporating hierarchical and semantic information, demonstrate substantially improved structure with clearer, more compact, category-specific clusters, reflecting a more clinically coherent representation. 
The legend represents the first character of ICD chapters~\citep{wikipediaICD10Wikipedia}, e.g., `R' corresponds to `Symptoms, signs and abnormal clinical and laboratory findings, not elsewhere classified'.}
\label{470697837357}
\vspace{5pt}
\end{figure*}

\vspace{-3.5mm} 
\subsection{Ablation Study}
To assess the individual contributions of HiSGT's components, we conduct an ablation study, systematically evaluating the impact of hierarchical embeddings, semantic embeddings, and consistency constraints on model performance (see Table~\ref{219470308897}). 
Starting with a base Transformer trained solely on medical codes (\texttt{Code Only}), we observe reasonable Unigram (0.956) and DimWise (0.964) scores, but it struggles with lower Bigram (0.898) and Seq Bigram (0.821) scores.
Introducing either hierarchical embeddings (\texttt{+ Hierarchical Embd}) or semantic embeddings ('\texttt{+Semantic Embd}) alone leads to gains, particularly improving Bigram scores to 0.914 and 0.913, respectively, and Seq Bigram to 0.845 and 0.836.
Combining both hierarchical and semantic embeddings (\texttt{+ Hierarchical \& Semantic Embds}) yields further improvements across all metrics, notably pushing Bigram to 0.927 and Seq Bigram to 0.859. This demonstrates the benefit of integrating both structural and semantic knowledge into the input representation.
Finally, adding the consistency regularization heads on top of both embeddings (\texttt{Full HiSGT}) provides the most significant boost. The full model achieves the best Bigram score (0.949, a substantial increase from 0.927 without consistency) and Seq Bigram score (0.879, up from 0.859). This highlights that both the enriched input embeddings and the consistency constraints guiding the generation process are crucial for HiSGT's superior fidelity. 

\vspace{-3.5mm}
\subsection{Interpretable Input Embeddings via t-SNE Visualizations}
To further evaluate the ability of our model to capture meaningful hierarchical and semantic information, we perform t-SNE visualizations of input embeddings in HiSGT and use HALO as a comparison. These visualizations help assess whether the model successfully groups medical codes into coherent clusters.
From Figure~\ref{470697837357}, we observe that HALO exhibits more scattered clusters, with many points overlapping across different ICD categories. This suggests that HALO's embedding space does not strongly enforce category-specific separation. In contrast, HiSGT demonstrates more defined and compact clusters, indicating that it captures category-level information more effectively.
The improved clustering structure also leads to a more interpretable embedding space.
This likely underpins the model's improved performance in capturing intra- and inter-visit code relationships (Table~\ref{017557960858}), ultimately contributing to the generation of more realistic synthetic EHR sequences.

\section{Conclusion}
In this study, we introduced HiSGT, a novel Hierarchy- and Semantics-Guided Transformer for synthetic EHR data generation.
By leveraging structured domain knowledge through hierarchical and semantic embeddings, HiSGT generates synthetic data that closely mirror real patient records and facilitates downstream clinical classification tasks. Importantly, the use of hierarchically and semantically meaningful code embeddings will contribute to more interpretable patient representations. 
While promising, the efficacy of HiSGT is partially constrained by the quality of the external knowledge sources (ontologies and pre-trained LMs). Additionally, assessing HiSGT's effectiveness across a broader range of downstream application, such as risk stratification or temporal event prediction, and extending HiSGT to handle non-ICD-coded data remain important directions for future work.

\clearpage
\bibliography{mybibfile}

\end{document}